\documentclass[runningheads]{llncs}

 \RequirePackage[table]{xcolor}
\usepackage[]{eccv}

\usepackage{eccvabbrv}

\usepackage{graphicx}
\usepackage{booktabs}
\usepackage{amsmath,amsfonts}
\usepackage{algorithmic}
\usepackage{algorithm}
\usepackage{array}
\usepackage{textcomp}
\usepackage{stfloats}
\usepackage{url}
\usepackage{verbatim}
\usepackage{cite}
\usepackage{bm}
\usepackage{makecell}
\usepackage{xcolor}
\usepackage{bbding}   
\usepackage{graphicx} 
\usepackage{multirow} 
\usepackage{pifont}    
\usepackage{pifont}  
\newcommand{\halfcheck}{%
\checkmark\kern-1.5ex\raisebox{0.7ex}{\rotatebox[origin=c]{135}{\textbf{--}}}%
}
\newcommand{\xmark}{\ding{55}}

\usepackage[accsupp]{axessibility}  

\usepackage{hyperref}
\usepackage{orcidlink}

\begin{document}

\title{SRGS: Super-Resolution 3D Gaussian Splatting} 

\titlerunning{SRGS: Super-Resolution 3D Gaussian Splatting}

\author{Xiang Feng\inst{1,2}
\and
Yongbo He\inst{3}
\and
Linxi Chen\inst{2} 
\and
Yan Yang\inst{1}
\and
Chengkai Wang\inst{1}
\and
Yifei Chen\inst{4}
\and
Yixuan Zhong\inst{5}
\and
Jiajun Ding\inst{1}$^{*}$ 
\and
Zhenzhong Kuang\inst{1}
\and
Xuefei Yin\inst{6}
\and
Yanming Zhu\inst{6}
}

\authorrunning{X. Feng et al.}

\institute{Hangzhou Dianzi University, Hangzhou, China \and ShanghaiTech University, Shanghai, China 
\and Zhejiang University, Hangzhou, China 
\and Tsinghua University, Beijing, China 
\and Xiangtan University, Hunan, China 
\and Griffith University, Brisbane, Australia}


\maketitle

\begin{abstract}
Low-resolution (LR) multi-view capture limits the fidelity of 3D Gaussian Splatting (3DGS). 3DGS super-resolution (SR) is therefore important, yet challenging because it must recover missing high-frequency details while enforcing cross-view geometric consistency. We revisit SRGS, a simple baseline that couples plug-in 2D SR priors with geometry-aware cross-view regularization, and observe that most subsequent advances follow the same paradigm, either strengthening prior injection, refining cross-view constraints, or modulating the objective. However, this shared structure is rarely formalized as a unified objective with explicit modules, limiting principled attribution of improvements and reusable design guidance. In this paper, we formalize SRGS as a unified modular framework that factorizes 3DGS SR into two components, prior injection and cross-view regularization, within a joint objective. This abstraction subsumes a broad family of recent methods as instantiations of the same recipe, enabling analysis beyond single-method innovation. Across five public benchmarks, we consolidate nine representative follow-up methods and trace reported improvements to specific modules and settings. Ablations disentangle the roles of priors and consistency, and stress tests under sparse-view input and challenging capture conditions characterize robustness. Overall, our study consolidates 3DGS SR into a coherent foundation and offers practical guidance for robust, comparable 3DGS SR methods. The code is available at {\url{https://github.com/XiangFeng66/SR-GS}}

\keywords{3D Gaussian Splatting \and Super Resolution \and Novel View Synthesis} 
\end{abstract}

\section{Introduction}
\label{sec:intro}
3D Gaussian Splatting (3DGS) \cite{3DGaussians} has recently emerged as a compelling representation for high-fidelity novel view synthesis due to its strong rendering quality and efficiency. However, in many practical settings, multi-view observations are captured at limited resolution because of sensor constraints, bandwidth, or legacy acquisition pipelines \cite{super-NeRF,wan2025s2gaussian,NeRF-SR}. This makes super-resolution (SR) 3DGS, that is, recovering a high-resolution (HR) 3DGS representation and HR renderings from low-resolution (LR) multi-view inputs, an important yet challenging problem. The difficulty stems from the fact that LR observations severely under-constrain high-frequency textures, while naive enhancement can easily introduce view-dependent artifacts that break multi-view coherence.

Our key insight that has shaped recent progress is that two complementary sources of information are required. First, 2D SR models provide strong priors for predicting high-frequency details from LR images. Second, the input multi-view observations inherently provide cross-view constraints that enforce coherence across viewpoints, thereby preventing view-dependent inconsistencies. Motivated by this observation, SRGS was introduced as the first simple and effective framework that systematically couples 2D SR priors with geometry-constrained cross-view consistency for learning HR 3DGS from LR multi-view capture.

Since SRGS, follow-up work has rapidly expanded. While these methods introduce different techniques and achieve improved performance, we observe that most methods can be interpreted as improved instantiations of the same underlying framework. Advances typically occur along three directions. One line strengthens 2D SR prior injection, for example, by adopting more powerful 2D SR backbones \cite{gaussiansr,zhang2025mvgsr,ko2024sequence,asthana2025splatsure}, injecting richer feature-level priors \cite{zhang2025mvgsr}, or improving the way 2D priors are distilled into 3DGS representation \cite{chen2025bridging}. A second line refines cross-view consistency regularization, for example, by designing stronger geometry-guided correspondence objectives \cite{ko2024sequence} or improving robustness to imperfect geometry and noise \cite{wan2025s2gaussian}. A third line changes {how the objective is applied}, for example, through term-specific reweighting \cite{xie2024supergs}, selective supervision \cite{asthana2025splatsure}, or masking strategies \cite{feng2025ie}. Some methods combine changes across multiple dimensions. Table \ref{tab:overview} provides a compact mapping of representative methods into this shared structure and summarizes their main improvement axes relative to SRGS.

Despite this progress, the field still lacks a clear, unified formulation that makes this shared structure explicit and disentangles what is shared across methods from what is novel, making it difficult to attribute reported gains to specific components, compare design choices at the module level, and derive reusable guidelines. This gap becomes more pronounced as the number of variants grows and improvements are introduced through different combinations of priors, consistency constraints, and objective modulation strategies. 


In this paper, we revisit SRGS and formalize it as a unified modular framework for 3DGS SR. We express 3DGS SR with a joint training objective that factorizes into two explicit modules, 2D SR prior injection and cross-view consistency regularization, and a well-defined interface for objective modulation through weighting and selection. This unified view subsumes a broad family of recent methods as different instantiations of the same framework, enabling principled analysis beyond single-method innovation. Across five public benchmarks, we consolidate nine representative follow-up methods and trace their reported improvements to specific modules and settings within the unified objective. We further conduct controlled ablations to isolate the respective contributions of priors and consistency, and perform stress tests under sparse-view input and challenging capture conditions to characterize robustness. Our study consolidates an emerging line of 3DGS SR research into a coherent and reproducible foundation and provides practical guidance for developing robust 3DGS SR methods.

\begin{table}[!h]
    \centering
    \caption{Representative 3DGS SR methods mapped to our unified SRGS framework. For each method, we indicate whether it conforms to the unified formulation and whether it instantiates the 2D SR prior module and the cross-view regularization module. The last column summarizes the primary modifications relative to SRGS.} 
    \label{tab:overview}
    \resizebox{\linewidth}{!}
    {
    \begin{tabular}{l@{\hspace{6pt}}c@{\hspace{6pt}}c@{\hspace{6pt}}c@{\hspace{2pt}}c}
    \toprule
    Methods & \makecell{Fits the Unified\\ Framework} & \makecell{2D SR \\ Prior Term} & \makecell{Input Geometry\\ Consistency Term} & \makecell{Main Improvement Axis}\\
    \midrule
    SRGS & \checkmark & \checkmark & \checkmark & - \\
    SuperGS (Arixv 2024) \cite{xie2024supergs} & \checkmark & \checkmark & \checkmark & weighted loss term\\
    GaussianSR (Arixv 2024) \cite{gaussiansr} & \halfcheck & \halfcheck & \checkmark & enhanced 2D SR prior; objective modulation \\
    Sequence Matters (AAAI 2025) \cite{ko2024sequence} & \checkmark & \checkmark & \checkmark & enhanced 2D SR prior \\
    MVGSR (Arixv 2025) \cite{zhang2025mvgsr} & \checkmark & \checkmark & \checkmark & feature-level prior distillation; additional loss term; selective supervision \\
    3DSR (ICCV 2025) \cite{chen2025bridging} & \xmark & \checkmark & \checkmark & new way of 2D SR prior injection\\ 
    S2Gaussian (CVPR 2025) \cite{wan2025s2gaussian} & \checkmark & \checkmark & \checkmark & selective supervision; additional loss term; objective modulation\\
    IE-SRGS (AAAI 2026) \cite{feng2025ie} & \checkmark & \checkmark & \checkmark & selective supervision; additional depth regluarization\\
    SplatSuRe (CVPR 2026) \cite{asthana2025splatsure} & \checkmark & \checkmark & \checkmark & selective supervision\\
    \bottomrule
    \end{tabular}
    }
\end{table}

Our contributions are as follows:
\begin{itemize}
    \item We formalize SRGS as a unified framework for 3DGS SR, with a modular objective that factorizes into 2D prior injection and geometry-constrained cross-view consistency regularization.
    \item We map nine latest representative 3DGS SR methods into the unified framework and analyze how their reported improvements arise from specific modules and settings on five public benchmarks.
    \item We present extensive ablations and stress tests to disentangle the roles of the 2D SR prior and the consistency regularization, assess the robustness of the unified framework under sparse views and challenging capture conditions, and derive practical design guidance for robust 3DGS SR systems.
\end{itemize}

\section{Related Work}
\subsection{3D Gaussian Splatting for High Fidelity Rendering}
We begin by reviewing the standard 3DGS reconstruction setting that motivates the 3DGS SR problem. 
3DGS \cite{3DGaussians,MipSplatting} has emerged as an efficient representation for high-fidelity novel view synthesis. It represents a scene as a set of anisotropic Gaussians with learnable attributes, and renders images through differentiable splatting, enabling fast optimization and real-time rendering while preserving fine appearance details. A large body of follow-up work extends the 3DGS pipeline with improved structure, scalability, or dynamic modeling \cite{scaffoldgs,pixelgs,4DGS,shi2025mmgs,sparse4dgs}. Most existing 3DGS reconstruction pipelines require sufficient HR multi-view observations, since the input image resolution and quality bound the supervision available for learning high-frequency textures and sharp edges. However, in practice, multi-view observations often have limited resolution due to sensor constraints, bandwidth limitations, or legacy acquisition protocols, which can significantly degrade the fidelity of the reconstructed 3DGS representation. 

This motivates 3DGS SR, which aims to recover an HR 3DGS representation from LR multi-view inputs. The problem is challenging because LR observations under-constrain high-frequency details, while naive enhancement can introduce view-dependent artifacts that violate multi-view coherence. Therefore, SRGS augments 3DGS reconstruction with additional priors and regularization, by injecting 2D SR priors to supply high-frequency details and by leveraging inherent geometry consistency in the input observations to impose cross-view consistency. 

\subsection{2D SR Priors for 3DGS Super-Resolution}
A natural approach to address the LR supervision bottleneck in 3DGS is to inject external 2D SR priors. 
Modern 2D SR models \cite{Swinir,EDSR,EDT,ESRGAN,FSRCNN,idm,hat,SR3} have demonstrated strong capability in predicting high-frequency details from LR inputs, driven by large-scale training data and advances in deep learning. Accordingly, many 3DGS SR methods incorporate 2D SR priors as supervision or regularization during optimization of an HR 3DGS representation. Common instantiations include: supervising rendered views with 2D SR pseudo-labels, distilling intermediate features or perceptual representations, and emphasizing high-frequency residuals. While 2D SR prior injection improves visual sharpness and texture fidelity, it also exposes a fundamental limitation of 2D SR priors. When applied independently per view, the injected details are not guaranteed to be consistent across viewpoints once lifted into a 3D representation. This often manifests as view-dependent textures and ambiguous renderings across viewpoints. These issues motivate explicit cross-view consistency regularization that couples views through geometry or multi-view correspondences.

\subsection{Cross-View Consistency Constraints}
Cross-view consistency is a fundamental principle in multi-view reconstruction \cite{chen2024mvsplat,zhang2025transplat,wang2025vggt} and novel view synthesis \cite{tang2023dreamgaussian,liu2024reconx,tang2024lgm} because it ties observations across viewpoints through geometry and discourages view-dependent artifacts. In the context of 3DGS SR, multi-view inputs induce geometric correspondences that can be exploited to regularize super-resolved details across views. Representative formulations include subpixel alignment-based consistency constraints and video SR-inspired multi-frame consistency regularization that enforces temporal or cross-view stability. Such constraints are effective at stabilizing details and improving multi-view coherence, but they can be sensitive to geometry noise, which may propagate errors through the consistency loss term. As a result, many recent methods co-design prior injection and cross-view consistency regularization, by objective modulation strategies such as selective supervision and confidence-weighted constraints. This trend highlights the need for a unified formulation that makes the shared structure explicit, isolates the contribution of each component, and enables principled comparisons and actionable design guidance.

\section{Problem Definition and Unified Framework}
\subsection{3DGS Preliminaries and Problem Setup}
We consider multi-view observations of a static scene. Let $\{\mathbf{I}^{v}_{\mathrm{LR}}\}_{v=1}^{V}$ denote the LR input images, with known camera parameters for each view $v$, as in standard 3DGS reconstruction pipelines. A 3DGS scene is represented by a set of $N$ anisotropic Gaussians, $\mathcal{G}=\{g_i\}_{i=1}^{N}$, where each Gaussian $g_i$ is parameterized by its 3D mean $\boldsymbol{\mu}_i \in \mathbb{R}^{3}$, covariance (or an equivalent scale and rotation parameterization) $\boldsymbol{\Sigma}_i$, opacity $\alpha_i$, and appearance attributes $\mathbf{c}_i$ (for example, color coefficients). Given $\mathcal{G}$ and camera $v$, a differentiable renderer $\mathcal{R}$ produces a rendered image at resolution scale $s$,
$
\hat{\mathbf{I}}^{v} = \mathcal{R}(\mathcal{G}, v; s).
$
In 3DGS SR, the goal is to recover an HR Gaussian representation $\mathcal{G}_{\mathrm{HR}}$ that enables HR renderings $\hat{\mathbf{I}}^{v}_{\mathrm{HR}}$, given only LR inputs $\mathbf{I}^{v}_{\mathrm{LR}}$. When real HR supervision is available, we denote the target images as $\mathbf{I}^{v}{\mathrm{HR}}$. When HR targets are provided by a 2D SR model for training, we denote them as $\tilde{\mathbf{I}}^{v}{\mathrm{HR}}$.

In the practical SR setting considered here, most 3DGS SR pipelines first increase representation capacity at high resolution, for example via Gaussian densification or splitting, and then rely on additional priors and regularization to guide the HR reconstruction. We treat this capacity initialization as a shared reconstruction step and focus on unifying how priors and multi-view consistency constraints enter the training objective.



\begin{figure*}[t]
\centering
\includegraphics[width=\textwidth]{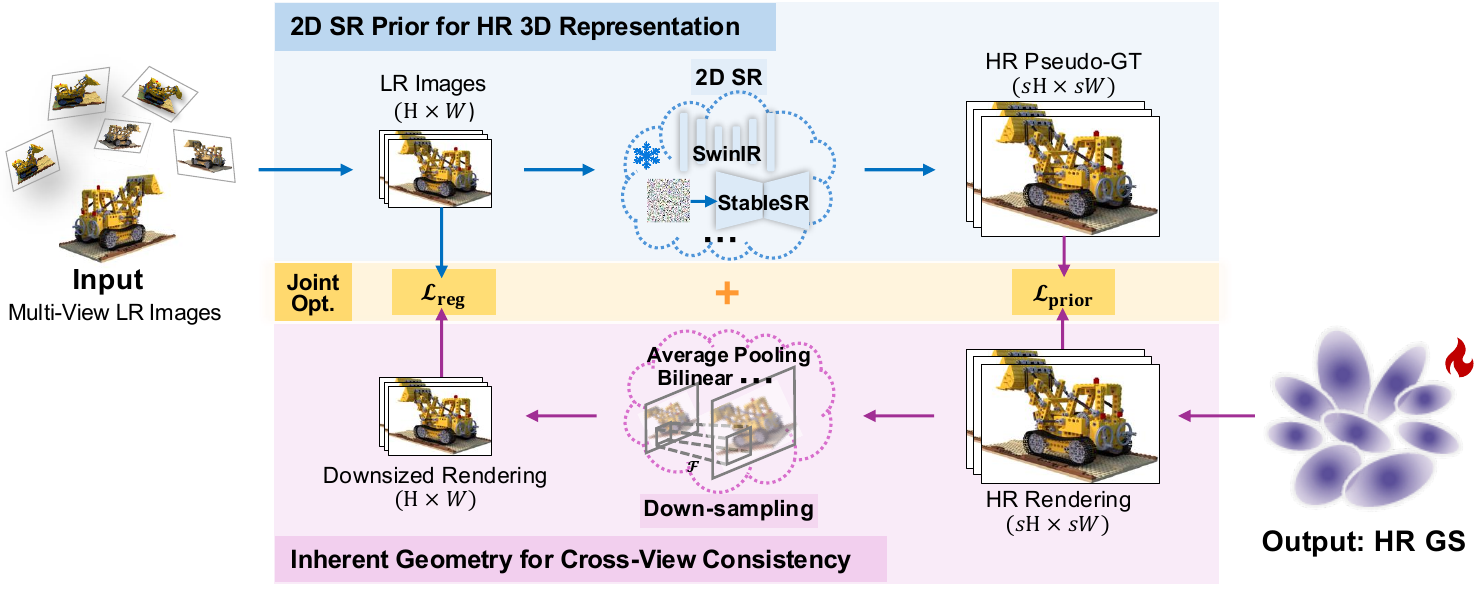}
\caption{Overview of the unified SRGS framework for 3DGS SR. Given multi-view LR inputs, a 2D SR prior branch produces pseudo HR targets to guide detail recovery, while a geometry-constrained branch enforces multi-view coherence via a render-and-downsample consistency constraint against the LR observations. The HR 3DGS representation is optimized jointly by combining the prior fidelity term $\mathcal{L}_{\mathrm{prior}}$ and the consistency regularizer $\mathcal{L}_{\mathrm{reg}}$.} 
\label{fig:framework}
\end{figure*}

\subsection{Unified Framework Overview}
Fig. \ref{fig:framework} illustrates the unified SRGS framework. Given LR multi-view inputs $\{\mathbf{I}^{v}_{\mathrm{LR}}\}_{v=1}^{V}$, we optimize an HR 3DGS representation $\mathcal{G}_{\mathrm{HR}}$ under two complementary sources of supervision.

\textbf{Prior injection module.}
A 2D SR model $\mathcal{M}$ provides view-wise prior signals that compensate for missing high-frequency content. In the simplest instantiation, $\mathcal{M}$ produces pseudo HR images:
\begin{equation}
{\mathbf{I}}^{v}_{\mathrm{HR}} = \mathcal{M}(\mathbf{I}^{v}_{\mathrm{LR}}),
\end{equation}
which act as targets for supervising HR 3DGS renderings. More generally, this 2D SR prior module can provide feature-level or perceptual guidance. Our unified formulation treats these variants as different instantiations of the same prior injection interface.

\textbf{Multi-view consistency regularization module.}
Multi-view observations are geometrically coupled because all views share a single 3DGS representation. We suppress view-dependent artifacts by enforcing a render-and-downsample consistency constraint between HR renderings and LR observations. Although applied per view, this term couples views through the shared $\mathcal{G}_{\mathrm{HR}}$ and known cameras, promoting coherent details across viewpoints. The regularizer can be instantiated in different forms, including subpixel consistency, correspondence-driven constraints, and selective supervision.


\subsection{Unified Objective Template}
We express 3DGS SR with a modular objective that combines a prior term and a geometry-constrained regularization term:
\begin{equation}
\mathcal{L}=\lambda_e \, \mathcal{L}_{\mathrm{prior}} + \left(1-\lambda_e\right)\mathcal{L}_{\mathrm{reg}},
\label{eq:unified_obj_template}
\end{equation}
where $\lambda_e$ controls the relative influence of the two modules. This template makes explicit the shared structure across recent 3DGS SR methods: improvements can arise from strengthening the prior module, refining the regularization module, or modulating how these terms are applied through strategies such as weighting, selection, and scheduling. Within our unified template, SRGS serves as a minimal yet effective instantiation, combining plug-in 2D SR prior supervision with a lightweight geometry-induced regularizer.

\textbf{2D SR prior term instantiation.}
In SRGS, we supervise HR renderings using pseudo HR targets from the 2D SR prior:
\begin{equation}
\mathcal{L}_{\mathrm{prior}}
=(1-\lambda_{\mathrm{tex}})\mathcal{L}_1\!\left(\tilde{\mathbf{I}}^{v}_{\mathrm{HR}},\hat{\mathbf{I}}^{v}(s_{\mathrm{HR}})\right)
+\lambda_{\mathrm{tex}}\mathcal{L}_{\mathrm{D\text{-}SSIM}}\!\left(\tilde{\mathbf{I}}^{v}_{\mathrm{HR}},\hat{\mathbf{I}}^{v}(s_{\mathrm{HR}})\right).
\label{eq:l_prior}
\end{equation}
This term injects high-frequency priors that are not directly recoverable from LR supervision alone.

\textbf{Multi-view consistency regularization term instantiation.}
To enforce HR renderings' geometry compatibility with LR inputs and suppress view-dependent artifacts, in SRGS, we match the LR observations with downsampled HR renderings. Let $\mathcal{F}$ denote a downsampling operator aligned with the LR formation process. We define
\begin{equation}
\mathcal{L}_{\mathrm{reg}}
=(1-\lambda_{\mathrm{cvc}})\mathcal{L}_1\!\left(\mathbf{I}_{\mathrm{LR}}^v,\mathcal{F}\!\left(\hat{\mathbf{I}}^{v}(s_{\mathrm{HR}})\right)\right)
+\lambda_{\mathrm{cvc}}\mathcal{L}_{\mathrm{D\text{-}SSIM}}\!\left(\mathbf{I}_{\mathrm{LR}}^v,\mathcal{F}\!\left(\hat{\mathbf{I}}^{v}(s_{\mathrm{HR}})\right)\right).
\label{eq:l_reg}
\end{equation}
This render-and-downsample constraint is intentionally lightweight, yet effective. It provides a geometry-induced coupling across views through the shared $\mathcal{G}_{\mathrm{HR}}$ and has served as a common starting point for subsequent studies.


Finally, our unified template also accommodates method-specific choices, including alternative penalty functions (for example, $\mathcal{L}_1$ versus $\mathcal{L}_2$), different prior injection signals (image-level versus feature-level), refined regularizers, and objective modulation strategies such as term-specific reweighting, selective supervision, and scheduling. We summarize these instantiations and their effects in the method mapping and analysis.

\section{Instantiations of the Framework}
Eq. \eqref{eq:unified_obj_template} makes explicit a shared structure in 3DGS SR: a \emph{2D SR prior injection} term that supplies high-frequency cues, and a \emph{geometry-constrained multi-view consistency regularization} term that enforces coherence through the shared 3DGS representation. Within this template, \textbf{SRGS is the minimal instantiation}, combining off-the-shelf 2D SR pseudo targets with a lightweight render-and-downsample regularizer. We show that a broad family of subsequent methods can be interpreted as \emph{non-exclusive} modifications along three dimensions:
(i) strengthening the prior module,
(ii) refining the multi-view regularizer,
and (iii) modulating the objective through strategies such as term-specific weighting, selective supervision, and masking. Table \ref{tab:overview} summarizes representative methods using this module-level view, turning the common recipe into a verifiable mapping rather than a narrative claim.

\subsection{2D SR Prior Injection Module}
SRGS instantiates the prior module with a plug-in 2D SR model that produces pseudo HR supervision for each input view. Follow-up methods that primarily improve this module keep the same training template but provide stronger or better aligned prior signals. This includes adopting more capable 2D SR backbones to produce higher-quality pseudo targets, injecting priors at richer feature levels instead of only image-level supervision, and improving how prior signals are distilled into 3DGS parameters to preserve fine textures. Representative examples are shown in Table \ref{tab:overview}: Sequence Matters \cite{ko2024sequence} and 3DSR \cite{chen2025bridging} mainly strengthen the prior injection pathway, whereas methods such as MVGSR \cite{zhang2025mvgsr} improve prior transfer through feature-level distillation and additional prior-related objectives, which still instantiate the same prior interface in Eq. \eqref{eq:unified_obj_template}. Importantly, these methods share the same core requirement: prior signals alone do not guarantee multi-view coherence once lifted into a shared 3D representation, which motivates complementary regularization.

\subsection{Cross-View Consistency Regularization Module}
The regularization module in SRGS is intentionally lightweight: it enforces compatibility with LR observations via a render-and-downsample constraint, which couples views through the shared $\mathcal{G}_{\mathrm{HR}}$ and known cameras. Subsequent work largely preserves this principle of geometry-induced coupling, but refines where and how the constraint is applied to improve stability and robustness. Common refinements include strengthening correspondence or alignment, restricting regularization to reliable regions, and reducing sensitivity to noise or imperfect pseudo-supervision. As summarized in Table~\ref{tab:overview}, methods such as SuperGS \cite{xie2024supergs}, S2Gaussian \cite{wan2025s2gaussian}, and SplatSuRe \cite{asthana2025splatsure} emphasize regularization-side refinements through weighted or selective constraints, while IE-SRGS \cite{feng2025ie} further augments coherence with auxiliary depth-related regularization. Although their implementations differ, they remain within the same unified template: they instantiate a geometry-constrained regularizer that suppresses view-dependent artifacts by coupling multiple views through shared 3D parameters.

\subsection{Objective Modulation and Auxiliary Terms}
Beyond changing the functional form of the two modules, many methods improve performance by modulating the effective influence of each term in the joint objective. This includes changing penalty functions (for example, $\ell_1$ versus $\ell_\text{MSE}$), introducing term-specific reweighting, and applying selective supervision or masking so that priors and consistency constraints focus on reliable pixels or regions. Several methods also introduce auxiliary losses (for example, $\ell_\text{TV}$), but these typically operate by reshaping either the prior term or the regularizer term, or by modulating their application rather than introducing a fundamentally new learning principle. In our unified view, such changes are naturally captured as objective modulation on top of the same two-module template in Eq.~\eqref{eq:unified_obj_template}.

\section{Experiments}
\subsection{Experimental Setup}
\label{sec:exp_setup}
\subsubsection{Datasets and Evaluation Protocol}
We evaluate on a diverse set of synthetic and real-world datasets commonly used for novel view synthesis. All experiments follow a strict LR supervision protocol: only bicubic-downsampled LR images are used for training, while evaluation is performed at the original resolution. We consider super-resolution factors of $\times4$ and $\times8$, depending on the dataset.

\textbf{NeRF Synthetic} \cite{nerf}.
NeRF Synthetic contains 8 synthetic scenes rendered at $800\times 800$ resolution. Following the standard split, we use 100 images per scene for training and 200 for testing. For $\times4$ SR, training images are downsampled to $200\times 200$, and evaluation is performed at $800\times 800$.

\textbf{Tanks and Temples} \cite{tanks}.
Tanks and Temples consists of real-world captures at $1920\times 1080$ resolution. We evaluate on a subset of 2 scenes. Following the standard protocol, we hold out $1/8$ of the views for testing and use the remaining views for training. For $\times4$ SR, training images are downsampled to $480\times 270$, and for $\times8$ SR to $240\times 135$, with evaluation at $1920\times 1080$.

\textbf{Mip-NeRF 360} \cite{mipnerf360}.
Mip-NeRF 360 includes 9 challenging real-world scenes with both indoor and outdoor environments. We follow the standard split, reserving $1/8$ of the views for testing. We report results under $\times4$ and $\times8$ SR settings by downsampling training images by the corresponding factors, while evaluating at the original resolution.

\textbf{DeepBlending} \cite{DeepBlending}.
DeepBlending contains 19 real-world scenes with diverse geometry and appearance. We adopt the standard split, using $1/8$ of the views for testing. Training images are downsampled by $\times4$ and $\times8$ for the corresponding SR settings, and evaluation is conducted at the original resolution.

\textbf{LLFF Dataset} \cite{mildenhall2019llff}.
LLFF consists of 8 forward-facing real-world scenes. Following the standard evaluation protocol, we select every eighth image for testing. To simulate sparse-view capture, we uniformly sample 3 views from the remaining images for training. For $\times4$ SR, training images are downsampled to $252\times 189$ and evaluated at $1008\times 756$.


\subsubsection{Methods and Comparison Protocol}
We compare SRGS with nine representative follow-up methods under consistent dataset splits, bicubic downsampling, test-view selection, and evaluation metrics. The compared methods include SuperGS \cite{xie2024supergs}, GaussianSR \cite{gaussiansr}, Sequence Matters \cite{ko2024sequence}, MVGSR \cite{zhang2025mvgsr}, 3DSR \cite{chen2025bridging}, S2Gaussian \cite{wan2025s2gaussian}, IE-SRGS \cite{feng2025ie}, and SplatSuRe \cite{asthana2025splatsure}. Interpreting these approaches as instantiations of our unified framework enables module-level attribution, namely to changes in prior injection, cross-view regularization, and objective modulation (for example reweighting, selective supervision, and scheduling). For each method, we follow the original design and recommended hyperparameters, and align evaluation settings to our protocol whenever applicable.

\subsubsection{Evaluation Metrics}
We evaluate HR novel-view quality by comparing rendered images with HR ground-truth views at the same camera poses. We report PSNR, SSIM \cite{SSIM}, and LPIPS (VGG) \cite{lpip}. Higher PSNR and SSIM indicate better fidelity, while lower LPIPS indicates better perceptual similarity. Metrics are computed on all test views and averaged per scene.

\subsubsection{Implementation Details}
Our implementation is based on the official open source 3DGS framework. We use the standard 3DGS training pipeline, including initialization, optimization schedule, and densification and pruning of Gaussian primitives. For each scene, we optimize the Gaussian representation for 30K iterations with the same base hyperparameters as 3DGS. For methods that require pseudo HR supervision, the pseudo targets are generated from LR training views using the method specified 2D SR prior. For SRGS, we use the pretrained SwinIR model \cite{Swinir} or StableSR model \cite{stableSR} to generate pseudo HR targets at the desired upscaling factor, and keep it fixed during training without scene-specific fine-tuning. The photometric penalty in both the texture supervision term and the consistency regularization term uses the same mixture weight as in 3DGS. Method-specific weights that balance the prior term and the consistency term follow the recommended settings in the corresponding papers. All experiments are conducted on a single NVIDIA RTX 4090 GPU.

\subsection{Main Results under a Unified Protocol}
\subsubsection{Quantitative comparison}
Tables~\ref{tab:main_results}, \ref{tab:main_results_sparse}, and \ref{tab:main_results_8scale} summarize quantitative results under the LR supervision protocol described in Sec.~\ref{sec:exp_setup}. We include the standard 3DGS baseline trained on LR inputs and SRGS, the minimal instantiation of our unified framework, instantiated with different 2D SR models for comparison. We then compare representative follow-up methods, which can be interpreted as modifying (non-exclusively) the 2D SR prior injection module, the multi-view consistency regularization module, or the way the joint objective is modulated through reweighting, selection, and scheduling. Some entries are missing because official implementations or pretrained models of these methods are not publicly available for all datasets and settings. When code is available, we reproduce results under the same splits, downsampling, and metrics. 

\begin{table}[t]
    \centering
    \caption{Quantitative comparison under the unified protocol for $4\times$ novel view super-resolution on NeRF Synthetic, Mip-NeRF 360, Tanks and Temples, and DeepBlending. Results are grouped by predictive-based and generative-based 2D SR priors. Missing entries indicate that an official implementation or a pretrained model is not publicly available for that dataset or setting.}
    \label{tab:main_results}
    \resizebox{\linewidth}{!}
    {
    \begin{tabular}{l@{\hspace{2pt}} ccc@{\hspace{6pt}} ccc@{\hspace{6pt}} ccc@{\hspace{6pt}} ccc@{\hspace{6pt}} ccc}
    \toprule[1pt]
    Methods & \multicolumn{3}{c}{NeRF Synthetic} & \multicolumn{3}{c}{Mip-NeRF 360} & \multicolumn{3}{c}{Tanks and Temples} & \multicolumn{3}{c}{DeepBlending} \\
    \cmidrule(lr){2-4} \cmidrule(lr){5-7} \cmidrule(lr){8-10} \cmidrule(lr){11-13} 
    Predictive-based & PSNR$\uparrow$ & SSIM$\uparrow$ & LPIPS$\downarrow$ & PSNR$\uparrow$ & SSIM$\uparrow$ & LPIPS$\downarrow$ & PSNR$\uparrow$ & SSIM$\uparrow$ & LPIPS$\downarrow$ & PSNR$\uparrow$ & SSIM$\uparrow$ & LPIPS$\downarrow$\\
    \toprule[1pt]
    3DGS Baseline \cite{3DGaussians}  &\cellcolor{green! 30} 21.77 & \cellcolor{green! 30}0.867 & \cellcolor{green! 30}0.104 & \cellcolor{green! 30}20.71 & \cellcolor{green! 30}0.619 & \cellcolor{green! 30}0.394 & \cellcolor{green! 30}19.62 & \cellcolor{green! 30}0.715 & \cellcolor{green! 30}0.337 & \cellcolor{green! 30}27.02 & \cellcolor{green! 30}0.851 & \cellcolor{green! 30}0.304 \\ \midrule
    SRGS (SwinIR) & 30.83 & 0.948 & 0.056 & 26.88 & 0.767 & 0.286 & 23.41 & 0.807 & 0.278 & 29.40 & 0.894 & 0.272\\
    SuperGS (Arixv 2024) \cite{xie2024supergs} & 30.89 & 0.949 & 0.056 & 27.12 & 0.768 & \textbf{0.262} & 21.19 & 0.695 & 0.364 & \textbf{29.77} & \textbf{0.899} & \textbf{0.268}\\
    SequenceMatters(AAAI 2025)\cite{ko2024sequence} & \textbf{31.41} & \textbf{0.952} & \textbf{0.054} & 27.02 & 0.775 & 0.277 & 23.43 & 0.808 & \textbf{0.274} & - & - & -\\
    MVGSR (Arixv 2025) \cite{zhang2025mvgsr} & 31.11 & 0.950 & 0.055 & - & - & - & 23.31 & \textbf{0.831} & 0.326 & - & - & - \\
    IE-SRGS (AAAI 2026) \cite{feng2025ie} & 30.97  & \textbf{0.952} & \textbf{0.054} & \textbf{27.15} & \textbf{0.779} & 0.278 & \textbf{23.52} & 0.810 & \textbf{0.274} & 29.63 & \textbf{0.899} & 0.271 \\
    \midrule
    Generative-based &  &  &  &  &  &  &  &  &  &  &  &\\
    SRGS (StableSR) & 28.45 & 0.923 & 0.093 & 25.92 & 0.734 & 0.317 & 22.58 & 0.757 & 0.282 & 28.23 & 0.861 & 0.317\\
    GaussianSR (Arixv 2024) \cite{gaussiansr} & 28.37  & 0.924 & 0.089 & 25.60 & 0.663 & 0.368 & - & - & - & 28.28 & \textbf{0.873} & 0.307\\
    3DSR (ICCV 2025) \cite{chen2025bridging} & - & - &   - & 26.10 & \textbf{0.746} & \textbf{0.222} & - & - & - & - & - & -\\ 
    SplatSuRe (CVPR 2026) \cite{asthana2025splatsure} & \textbf{28.94} & \textbf{0.932} & \textbf{0.083} & \textbf{26.34} & 0.740 & 0.323 & 22.95 & 0.770 & 0.267  &  \textbf{29.01} & 0.872 & \textbf{0.306} \\
    \bottomrule[1pt]
    \end{tabular}
    }
\end{table}

\paragraph{\textnormal{\textbf{Main $4\times$ results.}}}
Table~\ref{tab:main_results} reports $4\times$ novel view SR on NeRF Synthetic, Mip NeRF 360, Tanks and Temples, and DeepBlending. Compared with the LR-trained 3DGS baseline, SRGS improves PSNR by a significantly large margin across all datasets, confirming that coupling a 2D SR prior with geometry-induced regularization addresses the key ill-posedness of 3DGS SR. On Tanks and Temples and DeepBlending, SRGS also provides consistent gains over the LR baseline, indicating that the same two module template remains effective on real captures with more complex geometry and appearance. 

Beyond SRGS, the strongest follow-up methods typically add incremental improvements by strengthening one module or by modulating the objective. For predictive-based priors, improvements over SRGS are modest but consistent on datasets where results are available, for example, through refined regularization or selective supervision. This trend supports the central claim of our unified view: SRGS captures the shared core structure with a minimal objective, while later designs mainly refine how prior signals are injected, how cross-view constraints are applied, and how reliability is handled.

\begin{table}[!t]
    \centering
    \caption{Quantitative comparison under the unified protocol in the sparse-view setting for $4\times$ super resolution on NeRF Synthetic, Mip-NeRF 360, and LLFF. This setting is primarily targeted by S2Gaussian. For a fair comparison, SRGS adopts the FSGS initialization to match the sparse-view training regime used by S2Gaussian.} 
    \label{tab:main_results_sparse}
    \resizebox{\linewidth}{!}
    {
    \begin{tabular}{l@{\hspace{16pt}} ccc@{\hspace{12pt}} ccc@{\hspace{12pt}} ccc@{\hspace{12pt}} ccc}
    \toprule[1pt]
    Methods & \multicolumn{3}{c}{NeRF Synthetic} & \multicolumn{3}{c}{Mip-NeRF 360} & \multicolumn{3}{c}{LLFF} \\
    \cmidrule(lr){2-4} \cmidrule(lr){5-7} \cmidrule(lr){8-10}  
    & PSNR$\uparrow$ & SSIM$\uparrow$ & LPIPS$\downarrow$ & PSNR$\uparrow$ & SSIM$\uparrow$ & LPIPS$\downarrow$ & PSNR$\uparrow$ & SSIM$\uparrow$ & LPIPS$\downarrow$ \\
    \toprule[1pt]
    3DGS Baseline  & 20.65 & 0.831 &  0.141 & 13.01 & 0.272 & 0.489 & 16.25 & 0.342 & 0.474  \\
    SRGS + FSGS \cite{zhu2023FSGS} & 22.68 & 0.856 & 0.114 & 19.49 & 0.590 & 0.175 & 19.83 & 0.513 & 0.448  \\
    S2Gaussian (CVPR 2025) \cite{wan2025s2gaussian} & \textbf{24.19} & \textbf{0.879} & \textbf{0.089} & \textbf{20.45} & \textbf{0.654} & \textbf{0.139} & \textbf{22.05} & \textbf{0.687} & \textbf{0.296}  \\
    \bottomrule[1pt]
    \end{tabular}
    }
\end{table}

\begin{table}[!t]
    \centering
    \caption{Quantitative comparison under the unified protocol for $8\times$ super resolution on Mip-NeRF 360, Tanks and Temples, and DeepBlending. Only SplatSuRe targets this challenging higher-scale setting.}
    \label{tab:main_results_8scale}
    \resizebox{\linewidth}{!}
    {
    \begin{tabular}{l@{\hspace{16pt}} ccc@{\hspace{12pt}} ccc@{\hspace{12pt}} ccc@{\hspace{12pt}} ccc}
    \toprule[1pt]
    Methods  & \multicolumn{3}{c}{Mip-NeRF 360} & \multicolumn{3}{c}{Tanks and Temples} & \multicolumn{3}{c}{DeepBlending} \\
    \cmidrule(lr){2-4} \cmidrule(lr){5-7} \cmidrule(lr){8-10} 
    & PSNR$\uparrow$ & SSIM$\uparrow$ & LPIPS$\downarrow$ & PSNR$\uparrow$ & SSIM$\uparrow$ & LPIPS$\downarrow$ & PSNR$\uparrow$ & SSIM$\uparrow$ & LPIPS$\downarrow$  \\
    \toprule[1pt]
    3DGS Baseline   & 18.17 & 0.509 &  0.491 & 16.40 & 0.537 & 0.473 & 24.50 & 0.783 & 0.404 \\
    SRGS (StableSR)   & 24.88 & 0.648 & 0.422 & 21.97 & 0.674 & 0.399 & 27.35 & 0.830 & 0.377\\
    SplatSuRe (CVPR 2026) \cite{asthana2025splatsure}  & \textbf{25.08} & \textbf{0.655} & \textbf{0.406} & \textbf{22.48} & \textbf{0.692} & \textbf{0.378} & \textbf{28.03} & \textbf{0.843} & \textbf{0.357} \\
    \bottomrule[1pt]
    \end{tabular}
    }
\end{table}

\begin{figure*}[t]
\centering
\includegraphics[width=\textwidth]{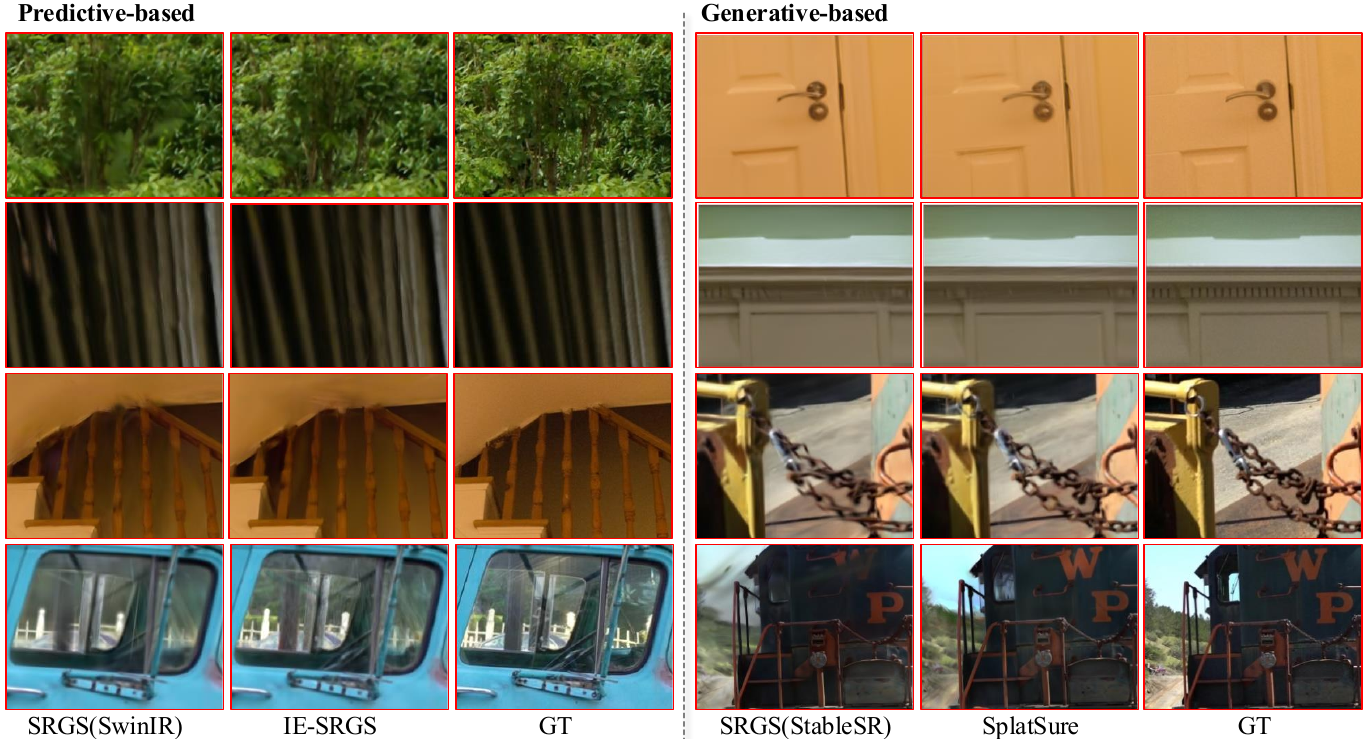}
\caption{Qualitative comparisons for $4\times$ novel-view SR. Left: predictive-based prior setting comparing SRGS(SwinIR) with IE-SRGS. Right: generative-based prior setting comparing SRGS(StableSR) with SplatSuRe.}
\label{fig:qualitative}
\end{figure*}

\paragraph{\textnormal{\textbf{Sparse view setting.}}}
Table~\ref{tab:main_results_sparse} evaluates a sparse view regime that is explicitly targeted by S2Gaussian. Training with only a few views substantially increases ambiguity, and the LR baseline degrades accordingly. Under the same sparse training setup, the SRGS-based framework remains effective and improves over the LR baseline on all datasets. S2Gaussian is the only method that targets this setting. It further improves upon the SRGS reference by introducing sparse-view-specific regularization and selection mechanisms, which aligns with our framework interpretation that objective modulation and selective constraints become increasingly important when view coverage is limited.

\paragraph{\textnormal{\textbf{Higher scale $8\times$ results.}}}
Table~\ref{tab:main_results_8scale} reports results in the more challenging $8\times$ SR on Mip-NeRF 360, Tanks and Temples, and DeepBlending. Even at this scale, SRGS remains a strong baseline and substantially improves over the LR-trained 3DGS baseline. Among methods with available results, SplatSuRe provides further gains by explicitly targeting this higher scale setting, indicating that scale-specific design choices can complement the same two-module foundation.

\subsubsection{Qualitative comparison}
Fig.~\ref{fig:qualitative} shows visual comparisons under predictive-based and generative-based priors. SRGS produces sharper structures than the LR baseline while maintaining cross-view coherence via consistency regularization. Follow-up methods that refine regularization or apply selective supervision further improve the stability of fine textures, reducing over-sharpened or view-dependent artifacts that can appear when prior signals are applied without sufficient reliability control. For generative-based priors, the qualitative results highlight the same pattern: improvements largely come from better balancing detail synthesis with cross-view coherence rather than introducing a fundamentally different training recipe.

\begin{table}[t]
\centering
\caption{Component ablation of the unified SRGS objective for $4\times$ 3DGS SR. We compare the LR-trained 3DGS baseline, adding only the 2D SR prior term, and the full objective that further includes the subpixel consistency regularizer. Results are reported across four datasets using two representative 2D priors, a predictive prior based on SwinIR and a generative prior based on StableSR.}
\resizebox{\linewidth}{!}{
\begin{tabular}{l ccc@{\hspace{6pt}} ccc c ccc@{\hspace{6pt}} ccc}
\toprule[1pt]
\multirow[c]{2}{*}{\makecell{SwinIR- \\based}} & \multicolumn{3}{c}{NeRF Synthetic} & \multicolumn{3}{c}{Mip-NeRF 360} & \multirow[c]{2}{*}{\makecell{StableSR- \\based}} & \multicolumn{3}{c}{Tanks and Temples} & \multicolumn{3}{c}{Deep Blending} \\
\cmidrule(lr){2-4} \cmidrule(lr){5-7} \cmidrule(lr){9-11} \cmidrule(lr){12-14}
 & PSNR$\uparrow$ & SSIM$\uparrow$ & LPIPS$\downarrow$ & PSNR$\uparrow$ & SSIM$\uparrow$ & LPIPS$\downarrow$ && PSNR$\uparrow$ & SSIM$\uparrow$ & LPIPS$\downarrow$ & PSNR$\uparrow$ & SSIM$\uparrow$ & LPIPS$\downarrow$ \\
\midrule[1pt]
3DGS baseline & 20.31 & 0.852 & 0.121 & 20.71 & 0.619 & 0.394 && 18.09 &0.649 &0.346 & 26.72 & 0.836 & 0.335 \\
+ 2D SR Prior & 30.38 & 0.945 & 0.059 & 26.66 & 0.762 & 0.301 && 21.71 & 0.732 & 0.302 & 27.16  & 0.845 & 0.325  \\
+ Consistency Regularizer & \textbf{30.83} & \textbf{0.948} & \textbf{0.056} & \textbf{26.88} & \textbf{0.767} & \textbf{0.286} && \textbf{22.58} & \textbf{0.757} & \textbf{0.282} & \textbf{28.23} & \textbf{0.861} & \textbf{0.317}  \\
\bottomrule[1pt]
\end{tabular} 
} 
\label{tab:ablation_components}
\end{table}

\subsection{Ablations of the Unified Framework}
\label{ablation}

\begin{figure*}[t]
\centering
\includegraphics[width=\textwidth]{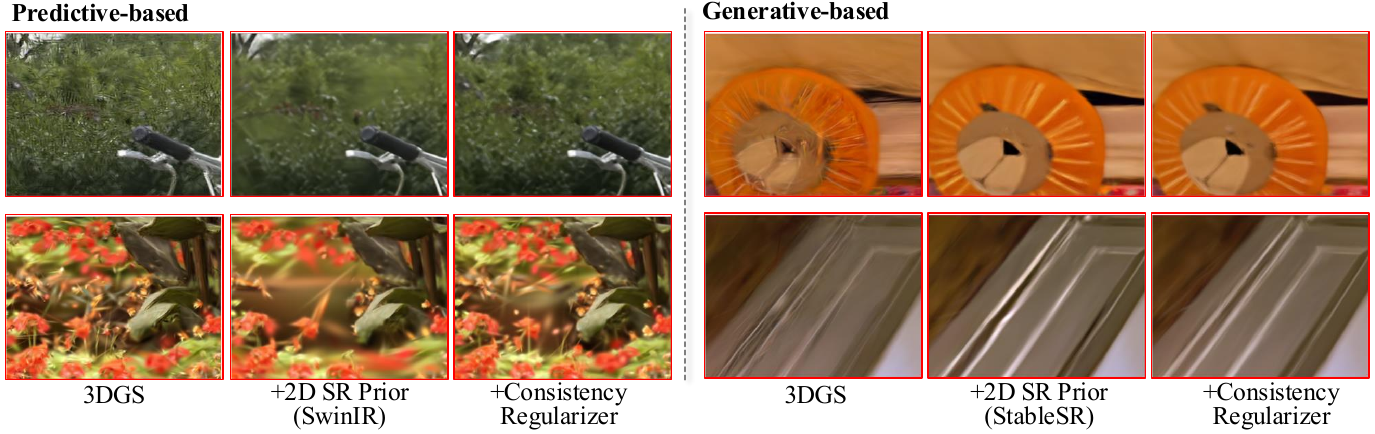}
\caption{Qualitative component ablation for $4\times$ 3DGS super-resolution under the LR supervision protocol. We compare the LR-trained 3DGS baseline (3DGS), adding the 2D SR prior only, and the full SRGS objective that further includes the consistency regularizer. Left: predictive prior (SwinIR). Right: generative prior (StableSR). }
\label{fig:ablation_vis}
\end{figure*}

\begin{figure}[t]  
    \centering  
    \vspace{-0.1cm}
    \includegraphics[width=0.7\textwidth]{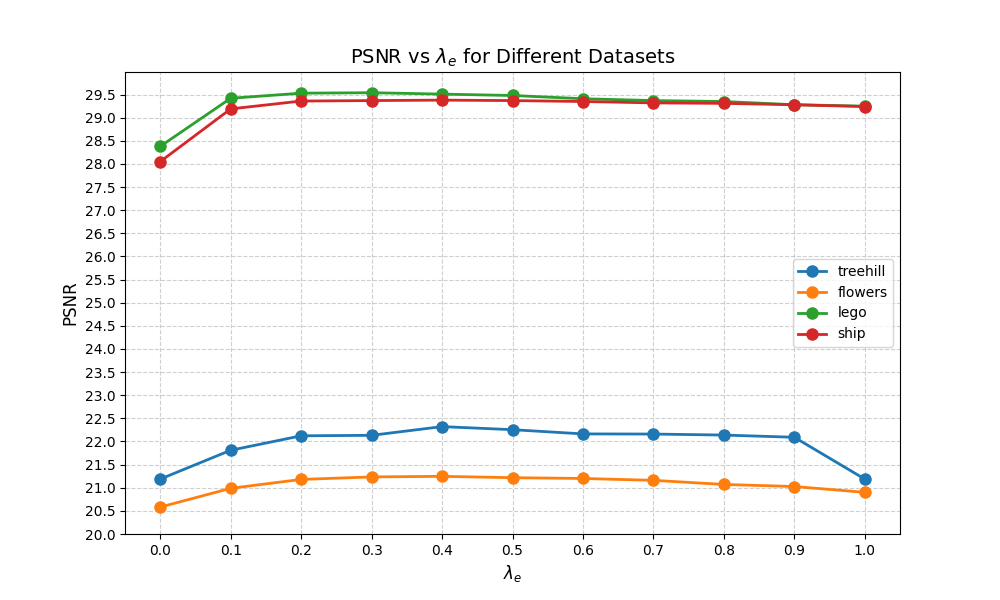} 
    \vspace{-0.2cm}
    \caption{Sensitivity to the balance weight $\lambda_e$ between the 2D SR prior and consistency regularization terms under $4\times$ SR. We report PSNR on four representative scenes.}
    \vspace{-0.1cm}
    \label{fig:lambda_sensitivity}  
\end{figure}

\subsubsection{Component analysis: prior injection vs. cross-view regularization}
Table~\ref{tab:ablation_components} quantifies the contributions of the two core modules in our unified objective. Starting from the LR-trained 3DGS baseline, adding only the 2D SR prior term yields a clear improvement across datasets, confirming that pseudo HR supervision is a key driver for recovering high-frequency content that is under-constrained by LR inputs. Adding the consistency regularizer on top further improves both distortion and perceptual metrics, indicating that geometry-constrained regularization complements the prior by enforcing multi-view coherence and suppressing view-dependent artifacts. This pattern holds for both predictive priors (SwinIR) and generative priors (StableSR), suggesting that the unified objective is robust to the specific choice of 2D SR model.

Fig.~\ref{fig:ablation_vis} provides qualitative evidence consistent with the quantitative trends. With only the 2D SR prior, textures become sharper but may exhibit unstable or view-dependent details. Adding the consistency regularizer reduces such artifacts and yields more coherent structures, especially around fine edges and repetitive patterns. Overall, the table and figure together support a module-level attribution: prior injection primarily drives detail recovery, while cross-view regularization stabilizes the recovered details and improves cross-view consistency.

\begin{table}[t]
\footnotesize
\centering
\caption{Stress tests under challenging conditions for $4\times$ SR. We report results on low-light and dynamic scenes using GS-DK and Deform-GS as task-specific backbones.}
\label{tab:stress_tests}
\resizebox{\linewidth}{!}{
\begin{tabular}{l@{\hspace{2pt}} c@{\hspace{6pt}}c@{\hspace{6pt}}c@{\hspace{12pt}} c@{\hspace{6pt}}c@{\hspace{6pt}}c}
\toprule
 Scene Conditions & \multicolumn{3}{c}{Low Light}  & \multicolumn{3}{c}{
Dynamic}   \\ 
\cmidrule(lr){2-4} \cmidrule(lr){5-7}
 & GS-DK \cite{gaussian-dk} Backbone  & +2D SR Prior  & +Regularizer & Deform-GS \cite{deformable3dgs}  Backbone   & +2D SR Prior & +Regularizer   \\ 
\midrule
PSNR$\uparrow$  &24.70 &25.15 &\textbf{25.55} &29.21 &39.41 &\textbf{39.68}\\
SSIM$\uparrow$ &0.790 &0.813 &\textbf{0.815} &0.961 &0.988 &\textbf{0.989}\\
LPIPS$\downarrow$ &0.167 &0.145 &\textbf{0.140} &0.037 &0.018 &\textbf{0.016}\\  
\bottomrule
\end{tabular}
}
\end{table}

\begin{figure}[t]  
    \centering  
    \includegraphics[width=0.8\textwidth]{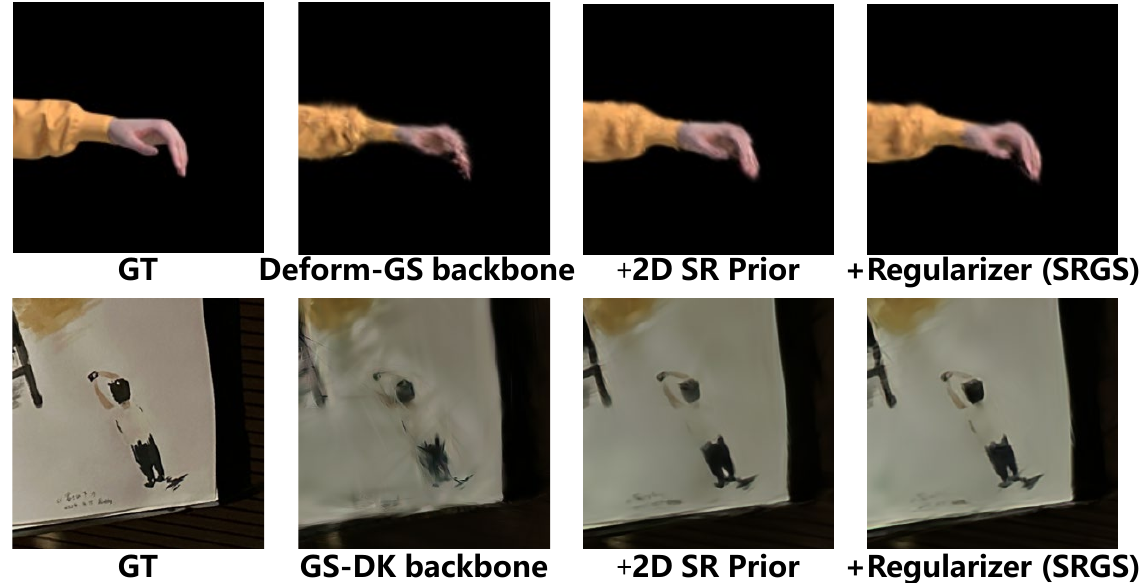} 
    \caption{Qualitative stress tests ($\times 4$ SR) under challenging capture conditions. Top: dynamic scene results using the Deform-GS backbone. Bottom: low-light scene results using the GS-DK backbone.}
    \vspace{-0.1cm}
    \label{fig:stress_vis}  
\end{figure}

\subsubsection{Sensitivity to the balance weight}
Fig.~\ref{fig:lambda_sensitivity} studies sensitivity to the balance weight $\lambda_e$ between the prior term and the consistency term. Across representative scenes, performance remains stable over a broad range of $\lambda_e$, with a mild optimum around the default value used throughout the paper. This behavior indicates that SRGS does not rely on delicate hyperparameter tuning. Instead, the unified objective provides a robust training recipe whose benefits persist under reasonable variations in the relative strength of the two modules.

\subsubsection{Stress tests under challenging conditions}
Table~\ref{tab:stress_tests} reports stress tests under challenging conditions, including low-light and dynamic scenes, using GS-DK and Deform-GS as task-specific backbones. Compared with LR-supervised backbones, adding the 2D SR prior consistently improves reconstruction quality, and further adding the consistency regularizer yields additional gains in both fidelity and perceptual similarity. Fig.~\ref{fig:stress_vis} shows the same trend qualitatively: the prior term sharpens details, while the full SRGS objective with regularization term further reduces unstable details and improves coherence. These results demonstrate that the SRGS objective is not tied to a particular 3DGS backbone or to standard static well-lit settings. Instead, it serves as a simple and reliable foundation that remains effective when the reconstruction setting becomes harder.

\section{Conclusion}
We revisited 3DGS SR and formalized SRGS as a unified, modular framework that factorizes the problem into prior injection and geometry-constrained cross-view regularization within a joint objective. This unified view subsumes a broad family of recent methods as instantiations of the same design space, enabling module-level comparison and attribution of performance gains. Extensive evaluations on five public benchmarks, together with controlled ablations, sensitivity analysis, and stress tests, show that SRGS remains a simple yet strong baseline and that later improvements largely refine the same core modules or modulate their application. Overall, our study consolidates this emerging line of work into a coherent foundation and provides practical guidance for developing robust and comparable 3DGS SR methods.


%
%
\bibliographystyle{splncs04}
\bibliography{main}

\end{document}